\documentclass[11pt,a4paper]{article}
\usepackage[hyperref]{emnlp2020}
\usepackage{times}
\usepackage{latexsym}

\usepackage{amsmath}
\usepackage{amsfonts}
\usepackage{natbib}
\usepackage{graphicx}
\usepackage{xcolor}
\usepackage{tikz}
\usepackage{subfig}
\usepackage{soul}
\usepackage[parfill]{parskip}
\usepackage{booktabs}
\usepackage{multirow}
\usetikzlibrary{arrows}
\usepackage{bbm}
\usepackage{amssymb}
\usepackage{sectsty}
\usepackage[shortlabels]{enumitem}
\usepackage{booktabs}
\usepackage{multirow}
\usepackage{color, colortbl}
\usepackage{tabularx}
\usepackage{pgfkeys}
\usepackage{dcolumn}
\usepackage{calculator}
\usepackage{makecell}
\usepackage{xcolor}
\usepackage{comment}
\usepackage{chngcntr}
\usepackage[colorinlistoftodos,prependcaption,textsize=tiny]{todonotes}
\usepackage{algorithm} 
\usepackage{algorithmicx}
\usepackage[noend]{algpseudocode}
\usepackage{mathtools, nccmath}

\newcommand*\Funct[2]{\textsc{#1}(#2)}
\newcommand*\Let[2]{\State #1 $\gets$ #2}
\renewcommand{\Comment}[2][.5\linewidth]{%
  \leavevmode\hfill\makebox[#1][l]{{\tiny $\triangleright$}~#2}}
  
\usepackage{scalerel, graphicx,xparse}
\NewDocumentCommand\image{}{\scalerel*{\centering
\includegraphics{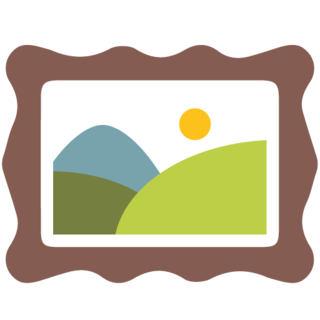}}{\vstretch{2}{X}}}

\definecolor{darkorange}{rgb}{0.8,0.32,0}
\definecolor{darkgreen}{rgb}{0.1,0.6,0.1}
\definecolor{cyan}{rgb}{0.88,1,1}
\definecolor{lightyellow}{rgb}{1.0, 1.0, 0.88}
\usepackage[export]{adjustbox}
\usetikzlibrary{arrows,positioning} 
\usetikzlibrary{shapes,snakes,automata,backgrounds,petri,shapes.geometric,calc,fit}
\tikzset{
    >=stealth',
    punkt/.style={
           rectangle,
           rounded corners,
           draw=black,
           text width=18em,
           minimum height=2em,
           text centered},
    dummy/.style={
           rectangle,
           rounded corners,
           draw=white,
           text width=15em,
           minimum height=2em,
           },
    pil/.style={
           ->,
           thick,
           shorten <=2pt,
           shorten >=2pt,}
}
\newcommand\mybox[3][l]{\setbox0\hbox{#2}\makebox[\the\wd0][#1]{#3}}

\tikzset{
  basic box/.style = {
    shape = rectangle,
    align = center,
    draw  = #1,
    fill  = #1!25,
    rounded corners},
  header node/.style = {
    minimum width = header nodes,
    font          = \strut\Large\ttfamily,
    text depth    = +0pt,
    fill          = white,
    draw},
  header/.style = {%
    inner ysep = +1.5em,
    append after command = {
      \pgfextra{\let\TikZlastnode\tikzlastnode}
      node [header node] (header-\TikZlastnode) at (\TikZlastnode.north) {#1}
      node [span = (\TikZlastnode)(header-\TikZlastnode)]
        at (fit bounding box) (h-\TikZlastnode) {}
    }
  },
  hv/.style = {to path = {-|(\tikztotarget)\tikztonodes}},
  vh/.style = {to path = {|-(\tikztotarget)\tikztonodes}},
  fat blue line/.style = {ultra thick, blue}
}

\usepackage{titlesec}
\usepackage{microtype}

\aclfinalcopy % Uncomment this line for the final submission
\newcommand{\jon}[1]{
}
\DeclareRobustCommand{\VAN}[3]{#2} % set up for citation

\newcolumntype{d}{D{.}{.}{-1}}

\usepackage[colorinlistoftodos,prependcaption,textsize=tiny]{todonotes}

\newcounter{mw}

\newcounter{af}

\newcounter{kl}
\newcommand{\eat}[1]{}
\title{\textsc{{C}ap{WAP}}: Captioning with a Purpose}

\author{Adam Fisch$^{1^*}$ \hspace{0.15em} Kenton Lee$^2$ \hspace{0.15em} Ming-Wei Chang$^2$ \hspace{0.15em} Jonathan H. Clark$^2$ \hspace{0.15em} Regina Barzilay$^1$ \\
$^1$Massachusetts Institute of Technology, $^2$Google Research \\
\texttt{\{fisch, regina\}@csail.mit.edu},\\
\texttt{\{kentonl, mingweichang, jhclark\}@google.com}}

\date{}

\begin{document}
\maketitle
\renewcommand{\thefootnote}{\fnsymbol{footnote}}
\footnotetext[1]{Work primarily completed while interning at Google.}
\renewcommand{\thefootnote}{\arabic{footnote}}
\begin{abstract}
%\jon{Title idea: Learning to Anticipate Visual Questions based on Information Needs: Image Captioning as a Latent Variable?}
%Different users care about different aspects of the world. The typical paradigm, however, %for building image captioning (image summarization) systems is to train models to imitate %one-size-fits-all reference captions. 

% TODO(jhclark): Rewrite to be even more clear that this is a task paper.
% TODO(jhclark): Say that we demonstrate that it's possible to build a system that does this.
The traditional image captioning task uses generic reference captions to provide textual information about images. 
Different user populations, however, will care about different visual aspects of images. 
In this paper, we propose a new task, {\bf Captioning with a Purpose} (\textsc{CapWAP}). Our goal is to develop systems that can be \emph{tailored} to be useful for the information needs of an intended  population, rather than merely provide generic information about an image. 
In this task, we use question-answer (QA) pairs---a natural expression of information need---from users, instead of reference captions, for both training and post-inference evaluation. We show that it is possible to use reinforcement learning to directly optimize for the intended information need, by rewarding outputs that allow a question answering model to provide correct answers to sampled user questions.
We convert several visual question answering datasets into \textsc{CapWAP} datasets, and demonstrate that under a variety of scenarios our purposeful captioning system learns to anticipate and fulfill specific information needs better than its generic counterparts, as measured by QA performance on user questions from unseen images, when using the caption alone as context.
\end{abstract}

\section{Introduction}
\label{sec:intro}

\begin{figure}[t!]
\centering
% Was 0.9\linewidth
\includegraphics[trim=60 75 10 60, clip,width=0.75\linewidth, frame=1pt]{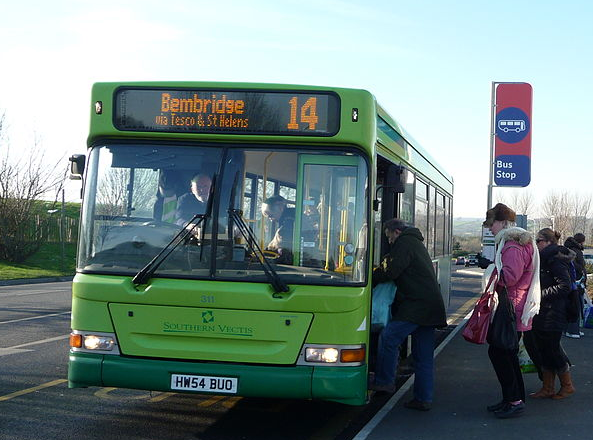}

\vspace{5pt}

\tabcolsep=0.2cm
\begin{center}
\footnotesize
\begin{tabularx}{7.5cm}{lll}
\toprule
\textbf{Task} &\textbf{Caption} & \textbf{Information Need} \\
\midrule
\\[-0.5em]
\makecell[l]{Captioning} & \makecell[l]{There is a green bus.} & \makecell[c]{\textit{(Unspecified)}} \\
\\[-0.5em]
\midrule
\\[-0.5em]
\makecell[l]{Visual QA} & \makecell[c]{\emph{(Unspecified)}} & \makecell[l]{Where's it headed?} \\
\\[-0.5em]
\midrule
\multirow{3}{*}{\textsc{CapWap}} & \multirow{3}{*}{\makecell[l]{At least three people \\ are boarding the \#14 \\ bus to Bembridge.}} & \makecell[l]{Which bus is this?} \\
& & \makecell[l]{Where's it headed?} \\
 &  & \makecell[l]{How many people \\ are boarding?} \\
\bottomrule
\end{tabularx}
\end{center}
\setlength{\belowcaptionskip}{-10pt}
\caption{The  informational purpose of generic captioning is not clearly defined, and VQA provides only reactionary information. The objective of the \textsc{CapWAP} task is ultimately to provide more informative captions that specifically \emph{anticipate} and \emph{satisfy} users' potential needs. In \textsc{CapWAP}, we use QA as an \textit{implicit} signal for information need: e.g., in the image above, a good caption that has been generated in advance should be able to be used to answer, \emph{Where is this bus headed?}}   %Only QA pairs, rather than captions, are available during training. \textsc{CapWAP} systems need to be tailored to the needs of a user population: the quality of captions is judged by their abilities to answer {\em future} questions that associated with test images.}
\vspace{-1pt}
\label{fig:main-example}
\end{figure}

\begin{table*}[!t]
\centering
\small
\tabcolsep=0.15cm
\renewcommand{\arraystretch}{1}
\begin{tabular}{llll}
\toprule
{\bf Task} & {\bf Training Data} & {\bf Prediction Function} & {\bf Evaluation}\\
\midrule
Captioning & \{\image, ref.caption\} & \{\image\} $\rightarrow$ pred.caption & \textsc{Similarity}(pred.caption, ref.caption)\\
Visual QA & \{\image, ref.question, ref.answer\} & \{\image, ref.question\} $\rightarrow$ pred.answer & \textsc{Accuracy}(pred.answer, ref.answer)\\
\cmidrule{1-4}
\textsc{CapWAP} & \{\image, ref.question, ref.answer\} & \{\image\} $\rightarrow$ pred.caption & \makecell[l]{\textsc{Accuracy}(\\
~~~~~~\textsc{QA}(ref.question, pred.caption),\\
~~~~~~ref.answer)} \\
\bottomrule
\end{tabular}
\caption{The \textsc{CapWAP} task combines elements of generic image captioning with visual question answering. Training consists of images paired with visual questions and answers. A  \textsc{CapWAP} model should directly {\em anticipate} user information needs by outputting \emph{captions} that can be used to answer {\em future} questions drawn from a distribution similar to the training data. Accordingly, the ``\textsc{QA}'' function represents the inference of an answer to a question using the generated caption as context. We approximate this with a strong automatic question answering model. }
\vspace{-5pt}
\label{tab:task-definition}
\end{table*}
% TODO: Can we change 'question' to 'questions'?
% TODO: Main metric: Informativeness / meeting human information need?
% TODO: Can we make it clear that it's possible that our captions can answer multiple questions?
% TODO: Suggestion from kentonl@: Say that our reference is a *set* of questions. 

The image captioning task typically selects for captions having high similarity with generic human references. While this task definition has driven much of the research in the field, the end-purpose of these captions is not always clearly articulated.
We argue that (1) generic annotations may not be representative of users' information needs, (2) user questions are a more natural way of articulating information needs, and (3) optimizing captions to provide correct answers to those questions allows training to focus on information need.
For example, in the VizWiz mobile application~\cite{BighamVizWiz}, visually impaired users upload images from their everyday lives, along with questions about them that need to be answered. These questions serve as a powerful signal for aspects of the image that they find important.

Consider the  image in Figure~\ref{fig:main-example}, where an annotator might provide a generic caption such as \textit{There is a green bus.} This may be used to answer: \textit{What color is the bus?} However, it would provide no utility to a user asking: \textit{Where is this bus headed?} In fact, examples from VizWiz demonstrate a clear disconnect between the type of information provided by today's  systems (e.g., arbitrary descriptions of entities and actions) versus what visually-impaired users need to know (e.g., fine-grained details to help make decisions).

Here, we propose an alternative framing for captioning: Captioning with a Purpose (\textsc{CapWAP}). We do not assume the existence of a \emph{universal} caption distribution. A good caption is highly subjective; different users will care about different aspects of a given image. Instead, we assume a distribution of visual question-answer pairs that are representative of population's information needs. Here we aim to map images to text that can serve as context to answer likely questions under this distribution. At test time, the goal is to anticipate similar user questions for a new image, and implicitly answer them {\em before} they even need to be asked.

We use image-question-answer triplets as supervision, and require the model to generate from the latent space of captions that provide contextual support for the answer (Table~\ref{tab:task-definition}). Within our task definition, any sampled caption that can be used to answer these questions is considered useful. Under this formulation, very different captions may be scored identically if they deliver the same content---regardless of word choice. Note that this is different from either standard visual question answering (VQA) or query-focused summarization: the target questions are not available prior to generation; at test time, they are used \emph{only} for evaluation.
%\mw{The last sentence is a bit hard to understand. Can we say the target questions are not available during generation --- and target questions only used for evaluation?}

Existing approaches cannot be readily applied in this setting,
% as we do \emph{not} assume the availability of 
as there are no gold reference captions for training---and off-the-shelf captioning systems transfer quite poorly
%\mw{It is not only an assumption. In fact, there no such data in the real world. So maybe we can say something stronger}
 (\S\ref{sec:generic-caption-evaluation}).
To address the new learning challenge that arises in \textsc{CapWAP}, we propose a novel model-in-the-loop reinforcement learning (RL) approach that acts as a strong baseline for this task. Our approach assumes a fixed question answering (QA) system that predicts an answer to a question using some input context. The captioning model receives a reward if it generates text which the QA system can use to predict the correct answer. 
Applying RL, however, is nontrivial. A na\"ive exploration of the caption generation space can lead to sparse rewards---resulting in long training times and disappointing quality. We show that our approach can be significantly improved by using a novel, synthetic pre-training routine to push the initial policy towards areas of high-reward.

We repurpose four VQA datasets for \textsc{CapWAP}: VQA~\cite{balanced_vqa_v2}, GQA~\cite{hudson2018gqa}, Visual7W~\cite{zhu2016cvpr}, and VizWiz~\cite{vizwiz-dataset}. These datasets range in style from synthetic QA pairs (GQA) to natural information-seeking questions asked by visually-impaired users (VizWiz). We find that our method produces significantly more informative captions with respect to the given questions (up to 3.8$\times$ exact match),  compared to models trained on generic captions from COCO~\cite{Lin2014MicrosoftCC}.
%Finally, in addition to increased efficiency, we find that using pre-training improves the fluency and image fidelity of the final policy.

Our key contributions are as follows:
\begin{enumerate}[leftmargin=*,noitemsep]
    \item We define a new task (\textsc{CapWAP}) that generates image captions for the \textit{purpose} of fulfilling specific information needs expressed by different target user populations.\vspace{10pt}
    \item We demonstrate that our information-need-driven model can generate much higher quality captions on this task than those of state-of-the-art traditional \emph{generic} captioning systems.\vspace{10pt}
    \item We propose a novel synthetic pre-training routine that greatly improves the performance of reinforcement learning under this new paradigm.
\end{enumerate}

\section{Related Work}
\label{sec:related}

Since the early days of the field, human-written references have been used for the supervised training and evaluation of text generation systems, including image captioning, summarization, and other related applications~\cite{edmundson-1969, lin-2003-automatic, ordonex2011im2text, vinyals2014neural}. Recently, researchers have begun to consider a multitude of different objectives for reference comparison~\cite{bohm-2019-better,gao2019reward}, or even parametric regressions trained on human judgements~\cite{louis-2013-automatically, peyrard-2018-objective}. Though diverse in approach, each ultimately relies on designing a robust general-purpose metric. In practice, engineering such a metric is challenging---if at all possible~\cite{sparck-jones-1994,sparck-jones-1999-factors}. Here we take a more empirical approach by relying on the information need expressed by users' questions. 

Many studies have observed that reference-trained captioning models suffer from systematic usability issues---including being rigid, neglecting relevant image aspects, and regurgitating frequent phrases~\cite{wang-cvae, Dai2017TowardsDA}. As a result, much effort has been focused on developing  secondary, corrective objectives---for instance, ``discriminability'' losses encouraging captions to be unique~\cite{dai-contrastive, liu-2018-showtell, Luo_2018_CVPR}. While these measures provide some fixes, they do not necessarily reflect user information needs---a central concept in \textsc{CapWAP}.
%; whereas the QA pairs used in our task are grounded in information need.

The idea of using QA for assessing information quality has been proposed in recent work for text summarization~\cite{arumae2019guiding, eyal-2019-apes,  scialom-2019-answers-unite}. The primary distinctions with our work are both the domain (images) and how questions are obtained---both of which impact the task objective and learning procedure. In this prior work, questions are generated programmatically~\cite[e.g., following][]{hermann-cnn-dailymail}.
Such ``questions'' may not necessarily reflect real user preferences. Our work focuses on QA not as just another method to improve standard reference-based metrics, but as a key, flexible way of formulating user information need---and as such we focus on challenging, \emph{real} QA datasets. Furthermore, we train on this signal, rather than rely on it solely for evaluation~\cite{wang2020asking}.

Efforts to leverage VQA resources to drive image captioning, and vice-versa, via variations of transfer learning, have also received extensive interest in recent years~\cite{li-etal-2018-tell, wu-etal-2019-generating, yang2019asking}. As opposed to optimizing metrics for specific  VQA or supervised captioning benchmarks, the primary focus in \textsc{CapWAP} is on modeling the target user population in order to anticipate the correct information-need. 

In a similar vein, VQA and textual QA resources have also been leveraged for active learning~\cite{shen2019active, Li2017LearningTD}, where the model learns to query its environment for information \emph{it} is uncertain about to help improve its performance on the given task. The key distinction with our work is the \emph{directionality} of the questions. In \textsc{CapWAP}, the model uses questions posed by the users to infer \emph{their} latent information need---which is a distinctly different, and quite challenging, setting.
\section{Problem Formulation}
\label{sec:formal}

\begin{figure*}[!t]
\centering
\small
\begin{tikzpicture}
  every node/.style={text centered,  minimum height=1.5em, minimum width=1.5cm,node distance=0pt},

  \tikzstyle{picture-style}=[rectangle,thick,draw=black!75]
  \tikzstyle{model-style}=[rectangle,thick,draw=blue!75,fill=blue!20,minimum size=6mm]
  \tikzstyle{data-style}=[rectangle,thick,draw=black!75,fill=black!20,minimum size=4mm]
  \tikzstyle{box-style}=[rectangle,thick,draw=black!75,fill=white!20,minimum size=6mm]

  \node[picture-style,left] (photo) {\includegraphics[scale=0.1]{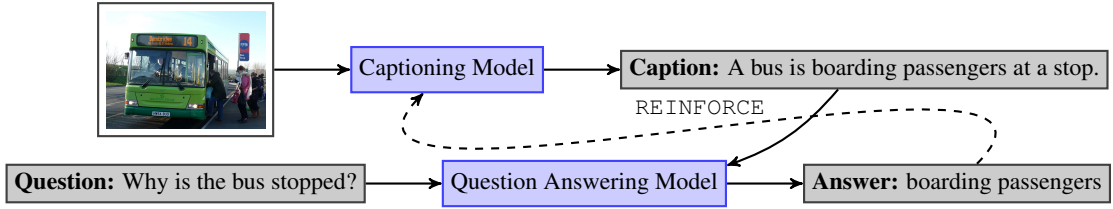}};
  \node[model-style, right = of photo] (caption-model) {Captioning Model};
  \node[data-style, right = of caption-model] (caption) {\textbf{Caption:} A bus is boarding passengers at a stop.};
    
  \node[data-style, below = 10pt of photo] (question) {\textbf{Question:} Why is the bus stopped?};
  \node[model-style, right = of question] (qa-model) {Question Answering Model};
  \node[data-style, right = of qa-model] (answer) {\textbf{Answer:} boarding passengers};
  
  %\begin{scope}[on background layer]
  %  \node[box-style, fit = (photo)(caption-model)(caption), header = Summarization System] (summarization-box) {};
  %\end{scope}
  
  %\begin{scope}[on background layer]
  %  \node[box-style, fit = (question)(qa-model)(answer), header = Discarded after training] (summarization-box) {};
  %\end{scope}
  
  \path[thick] (photo) edge[->] (caption-model) 
                      (caption-model) edge[->] (caption);
  \path[thick] (question) edge[->] (qa-model) 
                      (qa-model) edge[->] (answer);
  \path[bend left=15,thick] (caption) edge[->] (qa-model);
  \draw[thick,dashed, out=45, in=-135, ->] (answer) to node[above = 2pt of qa-model]{{\texttt{REINFORCE}}} (caption-model);
\end{tikzpicture}
\vspace{-14pt}

\caption{Overview of our proposed approach to the \textsc{CapWAP} task. The captioning model $G_\theta(\mathbf{y}|\mathbf{x})$ is learned using supervision from question-answer-image triples. Generated text that can be used to answer the question correctly, according to an extractive question answering model, is rewarded in our model-in-the-loop reinforcement learning framework. The questions, answers, and the question answering system are discarded after training.}
\vspace{-0pt}
\label{fig:framework}
\end{figure*}

% CapWAP task
% What is the input and what is the output?
% why is 

We begin by formulating the \textsc{CapWAP} task. 
In our setting, questions and answers are the only source of direct supervision assumed during training. At test time, the model is not given questions in advance, but rather must \emph{anticipate} the information need of the user, and generate captions that answer the forthcoming questions in expectation. %Below we formally define the task setting, information need, and question anticipation.

\textbf{Task Setting:} Given an image $\mathbf{x}$ the model must output a caption $\mathbf{y}$, such that $\mathbf{y}$ entails the answer $\mathbf{a}$ for a question-answer pair $(\mathbf{q}, \mathbf{a})$ sampled from some underlying distribution $\mathcal{D}$. Examples from  $\mathcal{D}$ are given during training, but are not known in advance by the generation model at test time. 

\textbf{Information Need:} We assume that the QA data from $\mathcal{D}$ is derived by the following process:
\begin{enumerate}[leftmargin=*, noitemsep]
\vspace{-2pt}
\item an image $\mathbf{x}$ is drawn from distribution $p(\mathbf{x})$;\vspace{4pt}
% \item a user $\mathbf{u}$ is drawn from distribution $p(\mathbf{u})$;\vspace{4pt}
\item a question-answer pair $(\mathbf{q}, \mathbf{a})$ targeting an informative detail of $\mathbf{x}$ perceived as important to a user in $\mathcal{D}$ is drawn from distribution $p(\mathbf{q}, \mathbf{a} | \mathbf{x})$.
\vspace{-2pt}
\end{enumerate}
The operating assumption is that the marginal distribution over $(\mathbf{q}, \mathbf{a})$ pairs represents the visual interests of the \emph{typical}  user. In other words, answers to common questions represent the type of information that is often considered important. This is comparable to \emph{content selection}~\citep{peyrard2019}.%~\citep{nenkova-2004-evaluating, peyrard2019}.

\textbf{Question Anticipation:} We do not assume the existence of a ``gold'' caption. Rather, the caption $\mathbf{y}$ is assumed to be a latent variable, and  $G_\theta(\mathbf{y} | \mathbf{x})$ is a stochastic generator that we must learn. A sample $\mathbf{y} \sim G_\theta(\mathbf{y} | \mathbf{x})$ should provide \emph{contextual support} for a new, randomly sampled question-answer pair. We estimate this using the accuracy of a pre-trained QA model $\mathcal{M}(\mathbf{q}, \mathbf{y})$, when using $\mathbf{y}$ as context for $\mathbf{q}$.  \textsc{CapWAP} requires maximizing the expectation:
\useshortskip
\begin{equation}
\label{eq:expectation}
\vspace{-1.5pt}
\arg\!\max_{\theta}\,\mathbb{E}_{G_\theta(\mathbf{y}| \mathbf{x})} \left[ \, \mathbb{E}_{p(\mathbf{q}, \mathbf{a}| \mathbf{x})} \left[\, \mathcal{R}(\mathbf{y} , \mathbf{q}, \mathbf{a}) \,\right] \, \right]
\end{equation}
where $\theta$ parameterizes  $G_\theta(\mathbf{y} | \mathbf{x})$, and we choose our reward to be $\mathcal{R}(\mathbf{y} , \mathbf{q}, \mathbf{a})$, any appropriate accuracy metric for comparing the output of $\mathcal{M}(\mathbf{q}, \mathbf{y})$ with $\mathbf{a}$ (expressed as \textsc{Accuracy} in Table~\ref{tab:task-definition}).

\textbf{\textsc{CapWAP} vs. Other Tasks:} 
Table~\ref{tab:task-definition} compares our setting to those of both standard (generic) captioning and visual question answering. Both standard captioning and \textsc{CapWAP} models output a single caption per image, but \textsc{CapWAP} does not compare to references. Both VQA and \textsc{CapWAP} models are trained and evaluated with QA data, but \textsc{CapWAP} does not provide the question prior to generation. VQA models output \emph{single answers}, whereas \textsc{CapWAP} models output anticipatory \emph{contexts}.
%\mw{We should add some sentences here for why CapWAP is needed if we can just use VQA}

\section{An Approach to \textsc{CapWAP}}

% TODO(jhclark): Add rhetorical questions to kick off each section?
% TODO(jhclark): (1) We use RL because...
% TODO(jhclark): (2) Optimization is hard because the model is rewarded only for correct answers, which is initially a rare event.
Given that we only have access to question-answer pairs during training, but not during inference, how can we learn a model for this task? Eq.~\ref{eq:expectation} naturally lends itself to a reinforcement learning (RL) framework where the model receives a reward $\mathbf{r}$ (e.g., $\mathbf{r} = \mathcal{R}(\mathbf{y} , \mathbf{q}, \mathbf{a})$) for each generated caption $\mathbf{y}$ and training QA pair $(\mathbf{q}, \mathbf{a})$. $G_\theta(\mathbf{y} | \mathbf{x})$ can be cast as a policy, and updated with policy gradients.

Optimizing such a policy, however, poses a technical challenge because the model is only rewarded for correct (or partially correct) answers, which is initially a rare event. Transferring $G_\theta(\mathbf{y} | \mathbf{x})$ from \emph{generic} captioning data can be a useful starting point. Our method then follows this  recipe:
\begin{enumerate}[leftmargin=*]
    \item Initialize $G_\theta(\mathbf{y} | \mathbf{x})$ using fully-supervised off-the-shelf captioning data, $(\tilde{\mathbf{x}}, \tilde{\mathbf{y}}) \sim \mathcal{D}_\text{generic}$; 
    \item Fine-tune $G_\theta(\mathbf{y} | \mathbf{x})$ using policy gradient on targeted visual QA data, $(\mathbf{x}, \mathbf{q}, \mathbf{a}) \sim \mathcal{D}_{\text{target}}$.
\end{enumerate}

In Sections~\ref{sec:modeling} and \ref{sec:learning} we detail our model for $G_\theta(\mathbf{y} | \mathbf{x})$, and the above training procedure.

Note that $\mathcal{D}_\text{generic}$ is assumed to be \emph{out-of-domain} for our intended captioning purpose, $\mathcal{D}_{\text{target}}$. Since we are interested in diverse user-generated questions and information needs, the generic captioning data can often diverge dramatically from our end goal. To improve transfer, in Section~\ref{sec:synthetic} we further develop a novel mechanism for automatically generating in-domain synthetic data that can be used as pre-training for guiding $G_\theta(\mathbf{y}|\mathbf{x})$ towards balanced areas of high reward in $\mathcal{D}_\text{target}$.

\subsection{Model Architecture}
\label{sec:modeling}

We briefly describe our base captioning model, which consists of a Faster R-CNN and Transformer-based encoder-decoder, following the sequence-to-sequence framework common in state-of-the-art image captioning systems \cite{Anderson2017up-down, vinyals2014neural, zhou2019vlp}.  See Appendix~\ref{app:model} for full technical  details.
Given an image $\mathbf{x}$, we first represent it as a sequence of detected object bounding box embeddings, computed from a pre-trained Faster R-CNN model \cite{Anderson2017up-down}. % Each region embedding is augmented with learned position, object detection confidence, and segment (i.e., ``image'' segment) embeddings.
We then generate caption word-pieces $\mathbf{y} = (y_1, \ldots, y_n)$ using a Transformer-based architecture~\cite{vaswani2017}.
% Each word-piece embedding is represented by the sum of its token, position, and segment (i.e., ``caption'' segment) embeddings. % following BERT~\cite{ devlin-2019-bert}.
%We condition on the image embeddings and decode auto-regressively---starting with the \texttt{[CLS]} token and terminating on the \texttt{[SEP]} token.

\subsection{Policy Training}
\label{sec:learning}
We describe our RL framework for training our captioning model using QA data. See Appendix~\ref{app:reinforce} for additional technical details, including hyper-parameter settings and optimization choices.

\textbf{Initialization:} We initialize $G_\theta(\mathbf{y} | \mathbf{x})$ using maximum likelihood estimation (MLE) on a corpus of out-of-domain generic captions $(\tilde{\mathbf{x}}, \tilde{\mathbf{y}})$, as common practice~\cite{Ranzato2015SequenceLT}. This warm-starts our policy with an initial set of grounded image concepts, albeit not necessarily the ones we ultimately care about. Given the generic reference $\tilde{\mathbf{y}} = (\tilde{y}_1, \ldots, \tilde{y}_n)$, we minimize the cross-entropy: 
\useshortskip
\begin{equation}
\label{eq:cross-entropy}
\vspace{-2pt}
\mathcal{L}_{XE}(\theta) = - \sum_{i=1}^{n} \log G_\theta(\tilde{y}_i \mid \tilde{\mathbf{x}}, \tilde{y}_{j < i})
\end{equation}
\useshortskip
\textbf{QA Model:} 
We implement the QA model $\mathcal{M}$ using a BERT$_{\text{LARGE}}$ extractive model fine-tuned on SQuAD 2.0~\cite{rajpurkar-2018-know}---which contains unanswerable questions. As an extractive model, $\mathcal{M}$ predicts a span $\mathbf{y}_{i \ldots j}$.
Important for our use-case,  $\mathcal{M}$ is both able to be accurate when predicting the answer $\mathbf{a}$ when $\mathbf{a}$ \emph{is} present in $\mathbf{y}$, and also able to abstain from answering when $\mathbf{a}$ is \emph{not} logically entailed (i.e., predict  ``no answer'').

\textbf{QA Reward:} We take $\mathcal{R}(\mathbf{y},\mathbf{q}, \mathbf{a})$ from Eq.~\ref{eq:expectation} as the F1 score of the predicted answer with the gold answer. We control for reward noise with a confidence threshold for predicting ``no answer.''
%instead of a span (to avoid potentially spurious rewards obtained by guessing or elimination.) 

\textbf{Policy Gradient:}
We use \texttt{REINFORCE} with a baseline \cite{williams1992} to compute the policy gradient $\nabla_\theta\mathcal{L}_{QA}(\theta)$ of the QA reward:

\useshortskip
\begin{equation}
\label{eq:pg}
\resizebox{.89\hsize}{!}{$
\hspace{-5pt}-\mathbb{E}_{G_\theta(\mathbf{y}| \mathbf{x})} \left[(\mathcal{R}(\mathbf{y},\mathbf{q}, \mathbf{a}) - b)\nabla_\theta\log G_\theta(\mathbf{y}| \mathbf{x})\right]
$}
\end{equation}
We take $b$ as $\mathcal{R}(\hat{\mathbf{y}}, \mathbf{q}, \mathbf{a})$, where $\hat{\mathbf{y}}$ is the $\arg\!\max$ (test-time prediction) of $G_\theta$, following the self-critical method of \citet{Rennie2016SelfCriticalST}.
\subsection{\hbox{Synthetic Policy Pre-Training}}
\label{sec:synthetic}

In the beginning of training, the generated captions typically do not correctly answer many questions, leading to almost no reward signal. More formally, the reward is sparse if the policy $G_\theta(\mathbf{y}|\mathbf{x})$ is not well-initialized. As a result, \texttt{REINFORCE} becomes extremely sample-inefficient.
When the target distribution is strikingly divergent from the one present in the generic captioning data---a key setting in this work---supervised pre-training on the out-of-domain data does not yield a usable initialization. As a substitute, we derive a method for generating a synthetic dataset of captions $\mathcal{D}_\text{synthetic}$ with high-reward as a form of guided policy search~\cite{levine-2013-gps}. The full method then consists of three stages that train on the three datasets: $\mathcal{D}_\text{generic} \rightarrow \mathcal{D}_\text{synthetic} \rightarrow \mathcal{D}_\text{target}$.

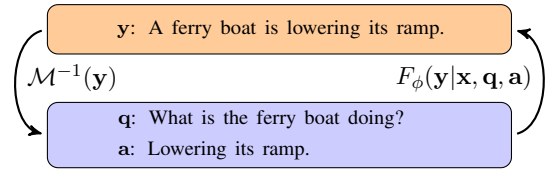
\begin{figure}[!t]
\centering
\begin{tikzpicture}[node distance=1cm, scale=.85, every node/.style={scale=.85}]
 \node[punkt, fill=orange!40] (text) {\small $\mathbf{y}$: A ferry boat is lowering its ramp.};
   \node[below=0.2cm of text] (dummy) {};
  \node[left=1.9cm of dummy] (f1) {$\mathcal{M}^{-1}(\mathbf{y})$};
  \node[right=1.3cm of dummy] (f2) {$F_\phi(\mathbf{y} | \mathbf{x}, \mathbf{q}, \mathbf{a})$};
  \node[punkt, below=0.2cm of dummy, fill=blue!20] (question) {\mybox[l]{\small $\mathbf{y}$: A ferry boat is lowering its ramp.}{\small$\mathbf{q}$: What is the ferry boat doing?} \linebreak \mybox[l]{\small $\mathbf{y}$: A ferry boat is lowering its ramp.}{\small $\mathbf{a}$: Lowering its ramp.}};
  \draw[pil, bend right=75]  (text.west) to (question.west);
  \draw[pil, bend right=75]  (question.east) to (text.east);
\end{tikzpicture}
\caption{A demonstration of reverse engineering the connections between question generation ($\mathcal{M}^{-1}$) and context generation ($F_\phi$). $\mathbf{x}$ is the image (not shown). See Algorithm~\ref{alg:learn} in Appendix~\ref{app:synthetic} for full details.}
\vspace{-12pt}
\label{fig:inversion}
\end{figure}

For the extractive QA model to possibly yield a positive reward, the answer must be a span of the caption. When the question and answer are known in advance, it is typically fairly simple to reverse engineer a candidate caption that meets this constraint (e.g., by inverting \emph{wh}-movement). Figure~\ref{fig:inversion} demonstrates this concept. If we have an auxiliary model $F_\phi(\mathbf{y} | \mathbf{x}, \mathbf{q}, \mathbf{a})$ that can automate this reverse engineering step, we can synthetically generate captions to use for pre-training, as in Eq.~\ref{eq:cross-entropy}.\footnote{Methods for constrained decoding \citep[][\emph{inter alia}]{Anderson2016GuidedOV, hokamp-liu-2017-lexically} that enforce $\mathbf{a} \subseteq \mathbf{y}$ are related, yet complementary, and can be incorporated into any $F_\phi$. It is more important to ensure that not only is the answer contained in the caption, but also that it is logically supported.}

\textbf{QA Conditional Model:} Motivated by this, we learn $F_\phi(\mathbf{y}|\mathbf{x}, \mathbf{q}, \mathbf{a})$ by explicitly conditioning on QA pairs when generating a caption that supports the answer span by design.
Concretely, we include the word-pieces of the question $\mathbf{q} = (q_1, \ldots, q_l)$ and answer $\mathbf{a} = (a_1, \ldots, a_m)$ as inputs when decoding $\mathbf{y}$, while $\mathbf{y}$ satisfies $\mathcal{M}(\mathbf{y}, \mathbf{q}) = \mathbf{a}$.

How do we train $F_\phi(\mathbf{y}|\mathbf{x}, \mathbf{q}, \mathbf{a})$ effectively without access to any paired data $(\mathbf{x}, \mathbf{q}, \mathbf{a}, \mathbf{y})$? We create automatic $(\tilde{\mathbf{x}}, \hat{\mathbf{q}}, \hat{\mathbf{a}}, \tilde{\mathbf{y}})$ examples from the out-of-domain generic captioning data used in Section~\ref{sec:learning} by using the (text-based) 
question generation model $\mathcal{M}^{-1}$ of \citet{alberti-2019-synthetic}.
%, trained on the extractive part of SQuAD 2.0. 
At a high level, given a generic caption $\tilde{\mathbf{y}} = (\tilde{y}_1, \ldots, \tilde{y}_n)$, this inverse model
(1) picks an answer span $\hat{\mathbf{a}} \subseteq \tilde{\mathbf{y}}$, (2) generates a question $\hat{\mathbf{q}}$ following some inferred distribution $p(\mathbf{q} | \hat{\mathbf{a}}, \tilde{\mathbf{y}})$, and (3) confirms that the sample obeys ``round-trip filtering'', i.e., that the original QA model answers the synthetic example correctly ($\mathcal{M}(\hat{\mathbf{q}}, \tilde{\mathbf{y}}) = \hat{\mathbf{a}}$).
% Given the generic caption $\tilde{\mathbf{y}}$, we generate $(\hat{\mathbf{q}}, \hat{\mathbf{a}})$, while enforcing  $(\hat{\mathbf{q}}, \hat{\mathbf{a}})$ to have round-trip consistency (i.e., $\mathcal{M}(\tilde{\mathbf{y}}, \hat{\mathbf{q}}) = \hat{\mathbf{a}}$).
We then train the model for $F_\phi$ using conditional cross-entropy:
\useshortskip
\begin{equation}
\label{eq:conditional-cross-entropy}
\vspace{-2pt}
\mathcal{L}_{CXE}(\phi) = - \sum_{i=1}^{n} \log F_\phi(\tilde{y}_i \mid \tilde{\mathbf{x}}, \hat{\mathbf{q}}, \hat{\mathbf{a}}, \tilde{y}_{j < i})
\end{equation}
\useshortskip
\textbf{Synthetic Data Generation:}
After training, we transfer $F_\phi(\mathbf{y}|\mathbf{x}, \mathbf{q}, \mathbf{a})$ with fixed weights to generate reverse engineered captions $\hat{\mathbf{y}}$ using the true $(\mathbf{x}, \mathbf{q}, \mathbf{a}, \texttt{null})$ examples from our target QA datasets. For each example, we decode the top-$k$ captions using beam search, and keep those with $R(\hat{\mathbf{y}},\mathbf{q}, \mathbf{a}) \geq c$, where $c$ is a threshold (e.g., $c=1.0$ for an exact match F1 score). These examples paired with the high-scoring captions are used to create the synthetic captions dataset $\mathcal{D}_\text{synthetic}$. We then use $\mathcal{D}_\text{synthetic}$ as further weak supervision for initializing $G_\theta(\mathbf{y} | \mathbf{x})$, again following Eq.~\ref{eq:cross-entropy}. See Appendix~\ref{app:synthetic} for additional technical details.

\section{Experimental Setup}
\label{sec:eval-metrics}

\textbf{Evaluation:}
Our primary evaluation assumes a dataset of questions and answers about images. Conceptually, if the correct answers are supported by the generated caption in expectation, then we consider it to be sufficiently informative.\footnote{Note that traditional captioning metrics such as ROUGE, BLEU, and CIDEr rely on gold references, which are not available in our new setting (in fact, by our definition, there is no one ``gold'' caption). Thus, we cannot include them.}

\textbf{Automatic Evaluation:}
Our automatic proxy of informativeness utilizes the state-of-the-art extractive question answering model ($\mathcal{M}$) described in Section~\ref{sec:learning} that is trained on SQuAD 2.0.\footnote{For evaluation we turn off the ``no answer'' option.}  $\mathcal{M}$ is applied to the given QA pair, with the generated caption as the ``context.'' We report EM, measuring exact match with the gold answer, and F1, measuring word overlap. If there are multiple answers, then we take the maximum score over all. %over all of them.

\textbf{Human Evaluation:}
For human evaluation we ask raters to judge whether a caption is less, equally, or more informative than another caption with respect to the question-answer pair. We also gather human ratings for two properties that are desirable regardless of the target audience: (1) fluency (whether the caption is grammatical and coherent) and (2) fidelity (whether the caption makes any false assertions regarding what is in the image).

\subsection{Datasets}
\label{sec:datasets}

We evaluate our method on four converted visual question answering datasets.
% Here we describe the data and pre-processing used for both pre-training and QA RL.\footnote{We distiguish these dataset splits with \emph{``Cap''} prefixes.} %For all datasets, we split text into uncased word-pieces using the vocabulary from \citet{devlin-2019-bert}.
We filter questions that are unaswerable, or have `yes/no' or non-alphabetic answers.\footnote{Numerical answers that are written out (e.g., \emph{two} vs. \emph{2}) are not disqualified. This requirement simplifies evaluation.} Appendix~\ref{app:datasets} gives additional size, splitting, and  pre-processing details.

\textbf{COCO}~\cite{Lin2014MicrosoftCC}: For all experiments, we use COCO as the source of out-of-domain generic captions for pre-training. COCO contains images covering 80 object categories and various scenes gathered from Flickr, paired with five human-written reference captions. 
% We use the `Karpathy' splits \cite{karpathy}.
%The annotators were prompted to describe thethe scene in at least eight words.

\textbf{CapVQA}~\cite{balanced_vqa_v2}: VQA v2.0 originally contains questions written by crowd-workers where the prompt was to write queries that are easy for humans to answer, but challenging for a hypothetical robot that mainly knows only about objects. VQA is the only dataset we consider that fully covers the same images as COCO.
% We re-partition the questions by image into train, dev, and test sets using the `Karpathy' splits. %Questions focus mainly on high-level scenes as well as properties of objects. 

\textbf{CapGQA}~\cite{hudson2018gqa}: GQA contains challenging compositional questions derived from scene graphs of everyday images using various human-specified grammars.
% We use the ``balanced'' splits, but limit to 5K images each for new test and dev sets from the original dev set.

\textbf{CapVisual7W}~\cite{zhu2016cvpr}: Visual7W contains questions written by crowd-workers about objects that, in general, require richer and longer answers than those in VQA. We use only the ``telling'' split of the dataset (i.e., the questions that require open-ended natural language answers). %We do not use the provided object bounding boxes. %The Visual7W images are sourced from Visual Genome, and thus some (imperfectly) overlap with the `Karpathy' COCO images. We map Visual Genome IDs to COCO IDs and re-split the dataset to follow $\text{train} = \text{COCO}_{\text{train}} \cup \text{Visual7W}_\text{train}$,  $\text{dev} = (\text{COCO}_{\text{dev}} \cup \text{Visual7W}_\text{dev}) \setminus \text{train}$, and $\text{test} = (\text{COCO}_{\text{test}} \cup \text{Visual7W}_\text{test}) \setminus (\text{train} \cup \text{dev})$.

\textbf{CapVizWiz}~\cite{vizwiz-dataset}: VizWiz consists of natural visual questions asked by visually-impaired users of a mobile application who were seeking answers to their daily visual needs. Each question is answered by a remote assistant.
%We take the most common option as the gold answer, and filter out questions that were deemed ``unanswerable.''
% We combine all of the original data, and re-partition it into test and dev sets of  $\sim$2K images each, keeping the rest for training.

\begin{table*}[!t]
\centering
\small
\tabcolsep=0.35cm
\renewcommand{\arraystretch}{1.1}
\begin{tabular}{l|dddddddd}
\toprule
\multirow{2}{*}{\textbf{Model}} & \multicolumn{2}{c}{\textbf{CapVQA}}            & \multicolumn{2}{c}{\textbf{CapGQA}}            & \multicolumn{2}{c}{\textbf{CapVisual7W}}       & \multicolumn{2}{c}{\textbf{CapVizWiz}}\\
 & \multicolumn{1}{c}{EM}  & \multicolumn{1}{c}{F1}          & \multicolumn{1}{c}{{EM}}  & \multicolumn{1}{c}{F1} & \multicolumn{1}{c}{{EM}} & \multicolumn{1}{c}{F1}      & \multicolumn{1}{c}{{EM}} & \multicolumn{1}{c}{F1}\\
\midrule
Human reference (from COCO)  & 16.5 & 25.7 & \multicolumn{1}{c}{-} & \multicolumn{1}{c}{-} & \multicolumn{1}{c}{-} & \multicolumn{1}{c}{-} & \multicolumn{1}{c}{-} & \multicolumn{1}{c}{-}\\
\citet{Luo_2018_CVPR} & 12.0 & 20.1 & 9.6 & 13.9 & 6.5 & 11.8 & 4.7 & 11.8 \\
\citet{huang2019attention} & 16.0 & 25.0 & 9.9 & 14.9 & 6.9 & 14.0 & 6.0 & 13.4 \\
Our generic baseline: \textsc{MLE} & 16.8 &  25.2 & 8.0 &  11.1 & 6.9 &    13.2           & 4.9 & 12.6\\
\cmidrule{1-9}
Our \textsc{CapWAP} model: \textsc{RL} & 23.1 & 32.3  & 15.7 & 19.3 & \textbf{10}.\textbf{5} & \textbf{18}.\textbf{4} & \textbf{22}.\textbf{5} & \textbf{28}.\textbf{5}\\
Our \textsc{CapWAP} model: \textsc{RL + SYN} & \textbf{24}.\textbf{2} & \textbf{33}.\textbf{2}  & \textbf{16}.\textbf{6} & \textbf{19}.\textbf{8} & 9.2 & 15.4 & {19}.{5} & {27}.{8}   \\
\bottomrule
\end{tabular}
\caption{\emph{Does the proposed approach better fulfill information needs?} We show question answering test performance when applying an extractive question answering model on predicted captions (see Table~\ref{tab:task-definition}). Existing captioning models trained on generic references (rows 2-4)---or even the generic references themselves (row 1)---do not capture the information requested by different QA datasets. Applying our RL method for tailoring towards \textsc{CapWAP} (row 5) leads to more informative captions with respect to those questions (and by extension, for the assumed end-users). Adding synthetic pre-training data (\textsc{+ SYN}) improves results on several datasets (row 6).}
\vspace{-0pt}
\label{tab:main-results}
\end{table*}

\begin{table}[!h]
\centering
\small
\tabcolsep=0.35cm
\renewcommand{\arraystretch}{1}
\begin{tabular}{l|cc}
\toprule
\multicolumn{3}{c}{\textbf{A} = {Purposeful \sc (RL + SYN)} vs. \textbf{B} = {Generic}}\\
\midrule
\multirow{2}{*}{\textbf{Dataset}\hspace{1.5cm}}&\multicolumn{2}{c}{\textbf{Informativeness}}\\
&\multicolumn{1}{c}{$A > B$} &\multicolumn{1}{c}{$B > A$} \\
\midrule
\text{CapVQA}& $\textbf{27\%}$ & $20$\%\\
\text{CapGQA}&  $\textbf{31\%}$ & $20$\% \\
\text{CapVisual7W}& $\textbf{38\%}$ & $22$\%\\
\text{CapVizWiz}& $\textbf{37\%}$  & $20$\%\\
\bottomrule
\end{tabular}
\caption{\emph{Do raters think that the proposed approach provides more informative captions?}
Human evaluation of the informativeness of captions with respect to our QA datasets
%$A > B$ indicates the percentage of raters that found system $A$ \emph{strictly} more informative than system $B$ and vice versa.
agrees with our automatic evaluation---finding our model to have better information coverage than \textsc{MLE}, our baseline without QA rewards.}
\vspace{-5pt}
\label{tab:human-eval}
\end{table}

% \subsection{Implementation Details}
% We use a 6-layer Transformer with 512 hidden input units, 8 attention heads, and 2,048 hidden units in the intermediate feed-forward layer. Image region embeddings are computed using the pre-trained model of \citet{Anderson2017up-down}. 
%  For MLE pre-training, all examples are shuffled and divided into mini-batches of 256 examples each. For RL adaptation, we use a mini-batch size of 128. For both settings, we train for a maximum of 120K steps and choose the best model based on the dev set (using COCO CIDEr %~\cite{VedantamZP15}
%  for MLE; F1 for RL). For optimization, we use Adam \cite{kingma-2015-adam} with a linear warm-up and decay schedule. \adam{Add info about MLE multi-tasking.}

% During inference we use a beam size of $k = 3$ for final predictions, and a beam size of $k=16$ for the PG rollouts. For both settings we use a length penalty $\alpha$ of $0.6$ \cite{wu2016google}. All models are implemented in Tensorflow \cite{tensorflow2015-whitepaper}.

\subsection{Generic Captioning Models}
In addition to our baseline captioning model trained only to maximize the likelihood of COCO references (\textsc{MLE} in the tables), we compare to two state-of-the-art \emph{generic} image captioning methods (also trained on COCO data). \citet{huang2019attention} directly optimizes the CIDEr metric with policy gradients, while \citet{Luo_2018_CVPR} optimizes both CIDEr and a ``discrimination'' loss intended to encourage models to describe each image's uniquely identifying aspects. % Since the objectives of both of these methods are primarily based on reference captions, we expect their improvements on COCO to be largely orthogonal to our QA metrics.
These models are included in order to highlight the differences in applicability between off-the-shelf models trained for generic image captioning versus those for \textsc{CapWAP}.
\section{Results}
\label{sec:experiments}

%This section presents the results of our empirical evaluation of our \textsc{Qa2Cap} framework.

%\subsection{Experimental Setup}

%\jon{In the QA performance table, I \textit{think} the difference between the top rows and bottom rows is that the top are trained on a \textit{single} dataset's generic target captions while the bottom rows are trained on dataset-specific QA pairs. If so, we should probably call this out in the caption.}

In the following, we address several key research questions relating to our approach to \textsc{CapWAP}, and the broader assumptions, strengths, and limitations of using QA to drive the process.% Appendix~\ref{app:experiments} describes additional analysis and ablations.

\textbf{\mbox{Evaluation of Generic Captions:}}
\label{sec:generic-caption-evaluation}
We begin by empirically verifying our introductory claim that training on generic reference captions can poorly reflect the varying, user-specific information need. Table~\ref{tab:main-results} presents the results of the baseline generic captioning systems when evaluated in terms of how well the predicted captions support QA over different distributions. Though they are strong methods as measured on the COCO benchmark, unsurprisingly, they still fail to capture all the information necessary to answer diverse visual questions. Performance on CapVizWiz is exceptionally poor; the visually-impaired users ask for information strikingly different than what is represented in COCO.
The causes of this poor performance go beyond simple limitations in the current state-of-the-art models; the target references themselves are insufficient. For example, on CapVQA, where the images overlap with COCO and thus human captions are directly available, the average performance of these ``gold'' references is only slightly better---supporting our conjecture that good captions for one purpose are not necessarily good for another, even on the same images.

%\jon{Try to answer the header question as the last sentence of the section, referring to specific results: \textit{Comparing rows X and Y, we see that systems trained on generic reference captions have much worse quality than systems trained on QA data.} (Or is there a better/stronger way of phrasing this answer?)}
%\jon{This might also be a good place to explain why our models are better than the human reference. Unexplainted, that looks quite suspicious.}

\begin{table}[!t]
\centering
\small
\tabcolsep=0.1cm
\renewcommand{\arraystretch}{1}
\newcommand{\qacell}[4]{ % #1: EM, #2: F1, #3: Mean of dataset.
\SUBTRACT{#2}{#3}{\normalizedfone}
\ADD{\normalizedfone}{5}{\offsetfone}
\MULTIPLY{\offsetfone}{4.}{\scaledfone}
\GLOBALCOPY{\scaledfone}{\scaledfone}
\cellcolor{darkgreen!\scaledfone !white}#2
}
\newcommand{\vqacell}[2]{\qacell{#1}{#2}{26.8}{black}}
\newcommand{\gqacell}[2]{\qacell{#1}{#2}{13.92}{black}}
\newcommand{\vsevenwcell}[2]{\qacell{#1}{#2}{13.68}{black}}
\newcommand{\vizwizcell}[2]{\qacell{#1}{#2}{17.86}{black}}
\begin{tabular}{l|dddd}
\cmidrule[\heavyrulewidth]{1-5}
 & \multicolumn{4}{c}{\textbf{Evaluation}} \\
\textbf{Train} & \multicolumn{1}{c}{CapVQA}            & \multicolumn{1}{c}{{CapGQA}}            & \multicolumn{1}{c}{{CapV7W}}       & \multicolumn{1}{c}{{CapVizWiz}}\\
\midrule
{CapVQA}             & \vqacell{17.7}{33.8}         & \gqacell{10.27}{14.3}                 & \vsevenwcell{9.05}{15.6}        & \vizwizcell{5.75}{16.4}      \\
{CapGQA}             & \vqacell{11.58}{24.8}          & \gqacell{16.43}{20.2}                 & \vsevenwcell{6.92}{12.6}         & \vizwizcell{3.95}{13.5}      \\
{CapV7W}        & \vqacell{12.32}{26.0}            & \gqacell{8.46}{11.4}                 & \vsevenwcell{9.02}{15.2}         & \vizwizcell{3.4}{14.0}      \\
{CapVizWiz}          & \vqacell{11.14}{23.9}            &\gqacell{8.9}{12.0}             & \vsevenwcell{6.71}{12.2}         & \vizwizcell{13.35}{31.3}      \\
\cmidrule{1-5}
{Generic} & \vqacell{11.71}{25.5}          & \gqacell{8.46}{11.7}               & \vsevenwcell{6.86}{12.8}      & \vizwizcell{3.7}{14.1}    \\
\bottomrule
\end{tabular}
\caption{\emph{Does the proposed approach tailor to specific information need?} We show transfer performance (\underline{dev} F1) of the \textsc{RL + SYN} policies learned on different QA datasets. The in-domain F1 peaks indicate that the model is producing distribution-specific captions.}
\vspace{-5pt}
\label{tab:cross-qa}
\end{table}

\newcommand{\imagecell}[1]{\raisebox{-.5\height}{#1}}

\begin{figure*}[t!]
\centering
%\vspace{5pt}
\tabcolsep=0.2cm
\begin{center}
\small
\begin{tabular}{llllll}
\toprule
\textbf{Dataset} &  \makecell{\textbf{Input}\\ \textbf{Image}} & \makecell[l]{\textbf{Generic Caption}\\\textbf{Output}} & \makecell[l]{\textbf{Purposeful Caption}\\\textbf{(RL + SYN)} \textbf{Output}}&
\makecell[l]{\textbf{Unseen} \\\textbf{Question}} & \makecell[l]{\textbf{Unseen}\\ \textbf{Answer}}
\\
% Height was 1.2cm
\midrule
\makecell[l]{{CapVQA}} &
\imagecell{\includegraphics[trim=60 65 10 50, clip, height=1cm, frame=0.5pt]{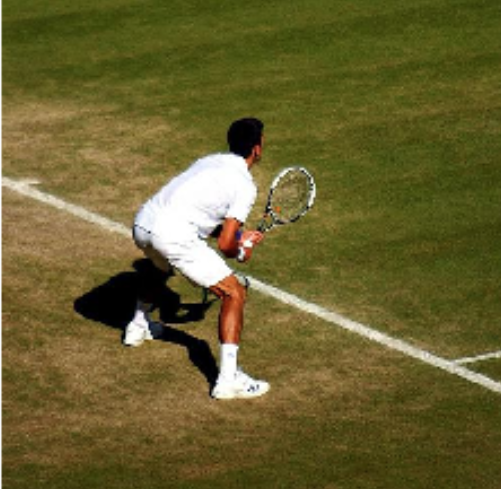}}&
\makecell[l]{a man playing tennis} & 
\makecell[l]{a man in white shirt and\\ white shorts playing a game\\ of tennis on a grass court}&
\makecell[l]{what color is\\ he wearing?} &
\makecell[l]{white shirt\\ and shorts} \\
\cmidrule{1-6}
\makecell[l]{{CapGQA}} &
\imagecell{\includegraphics[trim=60 0 50 10, clip, height=1cm, frame=0.5pt]{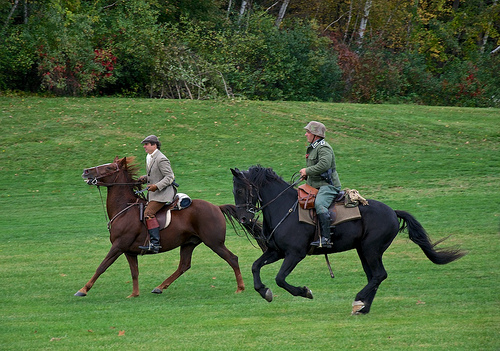}}&
\makecell[l]{a man riding on\\the back of a horse} & 
\makecell[l]{a man riding a horse next\\ to a woman on the right side}&
\makecell[l]{what color is the \\ horse the man is\\ to the right of?} &
\makecell[l]{brown}\\
\cmidrule{1-6}
\makecell[l]{{CapVisual7W}} &
\imagecell{\includegraphics[trim=75 20 75 10, clip, height=1cm, frame=0.5pt]{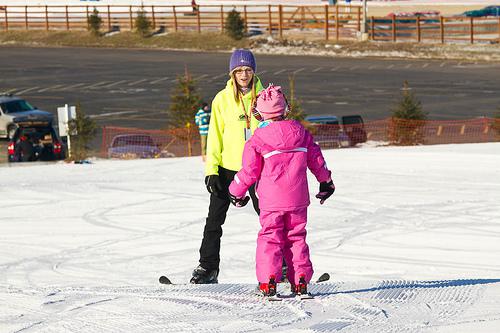}}&
\makecell[l]{a couple of people\\ that are standing\\ in the snow} & 
\makecell[l]{two people posing for a\\ picture taken at a ski slope}&
\makecell[l]{why are the kids\\ wearing coats ?} &
\makecell[l]{it is cold}\\
\cmidrule{1-6}
\makecell[l]{{CapVizWiz}} &
\imagecell{\includegraphics[trim=0 20 0 550, clip, height=1cm, frame=0.5pt]{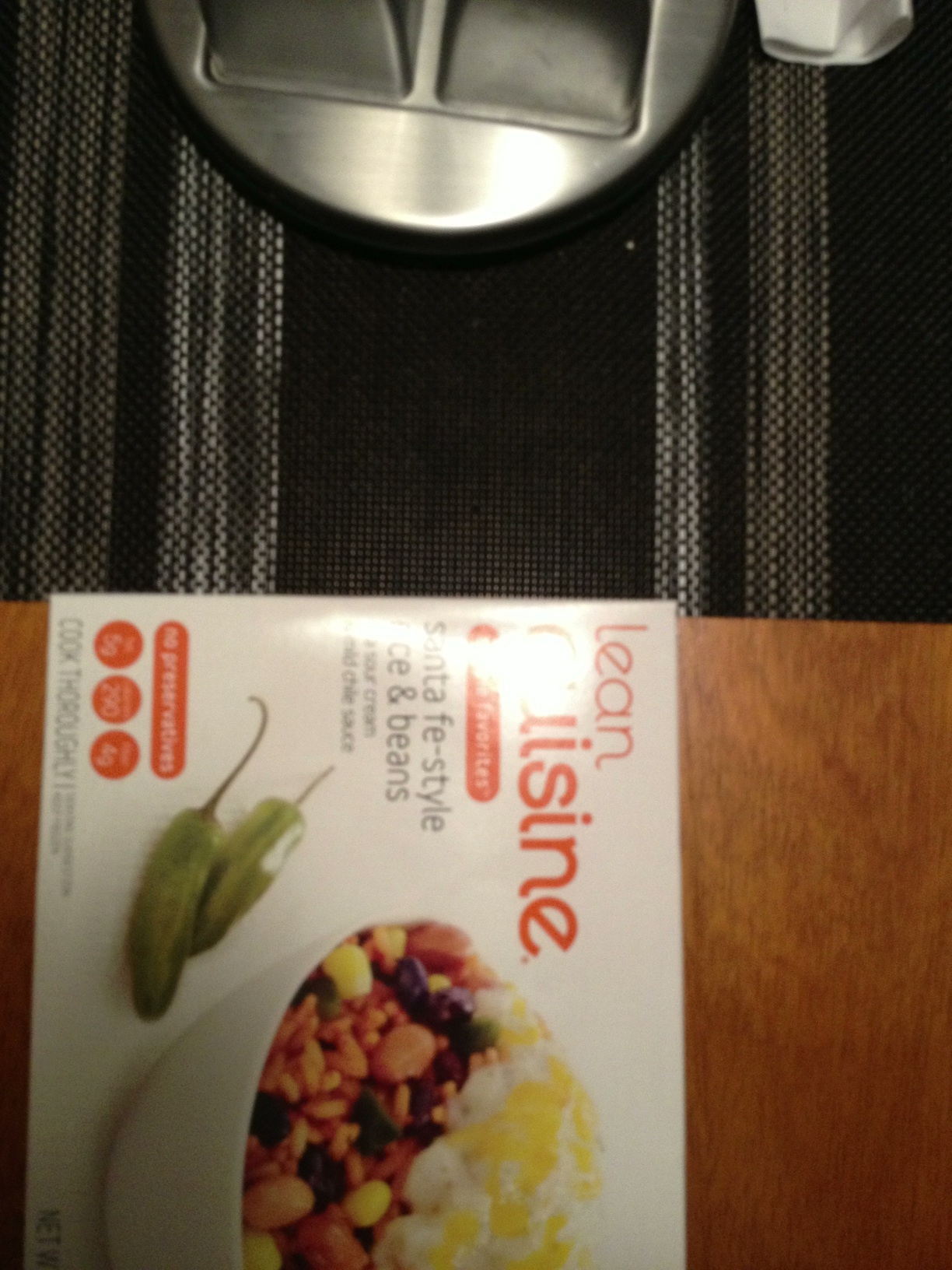}}&
\makecell[l]{a plate of food that\\ is on a table} & 
\makecell[l]{this is a picture of a sweet\\ corn frozen dinner}&
\makecell[l]{what kind of tv\\ dinner is this?} &
\makecell[l]{lean cuisine}\\
\bottomrule
\end{tabular}
\end{center}
\caption{Example outputs\protect\footnotemark~comparing our model trained for \textsc{CapWAP} (\textsc{RL + SYN}) with the baseline model trained on generic COCO references (\textsc{MLE}). These examples are representative of the way in which the various datasets ask about image content. Tendencies include colors for CapVQA, spatial relations for CapGQA, higher-level concepts in CapVisual7W, and OCR in CapVizWiz. Note that our approach to \textsc{CapWAP} does not (and likely cannot ever perfectly) anticipate \emph{all} unseen questions---but is distributionally closer in terms of content selection.}
\vspace{-5pt}
\label{fig:analysis-examples}
\end{figure*}

%\jon{I removed the hyphens from \textit{Information Need}. I would hyphenate noun compounds used as adjectives, but not when used as NPs.}
\textbf{\mbox{Adaptation to Information Need:}}
We next test the effectiveness of our proposed approach at tailoring captions to meet the specific information need stipulated by our datasets. Our results in Table~\ref{tab:main-results} demonstrate significant improvements by our QA-driven models (\textsc{RL} and \textsc{RL + SYN}) across all four datasets---achieving an average gain of 8.0 absolute F1. Notably, we improve by 7.5 EM over the average human caption on CapVQA, and by 16.5 EM  over the best generic model on CapVizWiz. Table~\ref{tab:cross-qa} further illustrates that the adaptation process is indeed tailored to the respective QA datasets.
The improvements on our automatic QA-based metrics (using the proxy model $\mathcal{M}$) also translate to human judgements. Table~\ref{tab:human-eval} presents the results of our human $A/B$ test of our proposed model vs. the \textsc{MLE} baseline. Relative to \textsc{MLE}, we find that our method is significantly more informative with respect to unseen QA pairs across all datasets. As expected, the largest improvements are on the datasets whose questions deviate significantly from the generic COCO content (e.g., CapVizWiz). 
%A relevant question to ask is if the captioning policies learned by \textsc{Qa2Cap} individually adapted to different information-need, or are the captions just becoming generally more informative?

\jon{Can you connect the 2.5X better claim from the intro in here?}

% \textbf{Disentanglement of Reward:}
% A motivating factor of the QA reward formulation is its robustness (in theory) to lexical realization---i.e., different captions may be scored identically if they deliver the same content. We qualitatively analyze the extent to which this is realized in our model in Figure~\ref{fig}.

\begin{table}[!t]
\centering
\small
\tabcolsep=0.15cm
\renewcommand{\arraystretch}{1.1}
\begin{tabular}{l|dd|dd}
\toprule
\multicolumn{5}{c}{\textbf{A} = {\sc RL + SYN} vs. \textbf{B} = {\sc RL}}\\
\midrule
\multirow{2}{*}{\textbf{Dataset}} &\multicolumn{2}{c}{\textbf{Fluency}}& \multicolumn{2}{|c}{\textbf{Fidelity}}\\
&\multicolumn{1}{c}{$A > B$} &\multicolumn{1}{c}{$B > A$} &\multicolumn{1}{|c}{$A > B$} &\multicolumn{1}{c}{$B > A$}\\
\midrule
\text{CapVQA}&\multicolumn{1}{c}{\textbf{57\%}} &\multicolumn{1}{c|}{26\%}&\multicolumn{1}{c}{\textbf{55\%}}& \multicolumn{1}{c}{29\%}\\
\text{CapGQA}&\multicolumn{1}{c}{\textbf{84\%}} &\multicolumn{1}{c|}{6\%}& \multicolumn{1}{c}{\textbf{76\%}}&\multicolumn{1}{c}{11\%}\\
\text{CapVisual7W}&\multicolumn{1}{c}{\textbf{69\%}} &\multicolumn{1}{c|}{19\%}& \multicolumn{1}{c}{\textbf{66\%}}&\multicolumn{1}{c}{22\%}\\
\text{CapVizWiz}&\multicolumn{1}{c}{\textbf{38\%}} &\multicolumn{1}{c|}{37\%}& \multicolumn{1}{c}{24\%}&\multicolumn{1}{c}{\textbf{47\%}}\\
\bottomrule
\end{tabular}
\caption{\emph{Does synthetic data improve secondary measures of caption quality?} Raters find that this strategy dramatically improves fluency and fidelity (\S\ref{sec:eval-metrics}) when compared to a model with only on-policy sampling.}
\vspace{-5pt}
\label{tab:syn-vs-rl}
\end{table}

\footnotetext{Examples are chosen to highlight model differences.}

\textbf{\mbox{Importance of Synthetic Pre-training:}}
A deficiency of QA-based rewards is that they neither explicitly enforce text fluency, nor penalize the system when content is produced that is either \emph{not} relevant or \emph{not} true. On the other hand, when reference captions are available, it is easy to learn a fluent language model. Table~\ref{tab:syn-vs-rl} shows that incorporating synthetic, guiding ``silver'' samples from our auxiliary QA conditional model $F_\phi(\mathbf{y} | \mathbf{x}, \mathbf{q}, \mathbf{a})$ (\S\ref{sec:synthetic}) to bridge the gap between $\mathcal{D}_\text{generic}$ and each considered $\mathcal{D}_\text{target}$ dramatically reduces the fluency and fidelity issues that arise from training solely with QA rewards. Ultimately, however, Table~\ref{tab:fluency-and-fidelity} shows that our model still suffers on these secondary metrics as compared to the reference-trained MLE baseline. This is a challenge that is shared with nearly all other comparable RL-based methods for text generation~\cite[e.g.,][\emph{et cetera}]{guo-2018-improving, paulus}.  Incorporating complementary fluency rewards (e.g., via pre-trained language model perplexity) is a valuable direction for future work.

\begin{table}[t!]
\centering
\small
\tabcolsep=0.2cm
\renewcommand{\arraystretch}{1}
\begin{tabular}{l|c|c}
\toprule
\multicolumn{3}{c}{\textbf{A} = {Purposeful \sc (RL + SYN)} vs. \textbf{B} = {Generic}}\\
\midrule
\multirow{2}{*}{\textbf{Dataset}} &\multicolumn{1}{c}{\textbf{Fluency}}& \multicolumn{1}{|c}{\textbf{Fidelity}}\\
&\multicolumn{1}{c}{$A \geq B$} &\multicolumn{1}{|c}{$A \geq B$} \\
\midrule
\text{CapVQA}& \multicolumn{1}{c}{37\%} &\multicolumn{1}{|c}{33\%} \\
\text{CapGQA}&  \multicolumn{1}{c}{30\%} &\multicolumn{1}{|c}{\textbf{57\%}} \\
\text{CapVisual7W} &\multicolumn{1}{c}{37\%} & \multicolumn{1}{|c}{\textbf{53\%}} \\
\text{CapVizWiz}& \multicolumn{1}{c}{34\%} &\multicolumn{1}{|c}{\textbf{72\%}} \\
\bottomrule
\end{tabular}
%\vspace{-0.1cm}
\caption{\emph{Do more informative captions come at a cost?}
% of fluency or fidelity (\S\ref{sec:eval_metrics})?}
Raters find that our tailored ``purposeful''  approach is less fluent more than half the time (left). Still, this system has greater or equal fidelity to the image content on most datasets (right). At a high level, while the system does give more \emph{relevant} information, it may do so in a less fluent way, a direction we leave for future work.}
\vspace{-5pt}
\label{tab:fluency-and-fidelity}
\end{table}

\begin{table*}[!t]
\centering
\small
\resizebox{1\textwidth}{!}{%
\renewcommand{\arraystretch}{1.2}
\begin{tabular}{@{}l|ddd|ddd|ddd|ddd@{}}
\toprule
\multirow{3}{*}{\textbf{Model}}          & \multicolumn{12}{c}{\textbf{Dataset}}                                                                                     \\
                                         & \multicolumn{3}{c}{CapVQA} & \multicolumn{3}{c}{CapGQA} & \multicolumn{3}{c}{CapVisual7W} & \multicolumn{3}{c}{CapVizWiz} \\
 &
  \multicolumn{1}{c}{$|\mathbf{y}|$} &
  \multicolumn{1}{c}{adj\%} &
  \multicolumn{1}{c}{$|V^2|$} &
  \multicolumn{1}{c}{$|\mathbf{y}|$} &
  \multicolumn{1}{c}{adj\%} &
  \multicolumn{1}{c}{$|V^2|$} &
  \multicolumn{1}{c}{$|\mathbf{y}|$} &
  \multicolumn{1}{c}{adj\%} &
  \multicolumn{1}{c}{$|V^2|$} &
  \multicolumn{1}{c}{$|\mathbf{y}|$} &
  \multicolumn{1}{c}{adj\%} &
  \multicolumn{1}{c}{$|V^2|$} \\ \midrule
Generic: MLE                   & 10.8     &  6.9   & 3.6   & 11.1     & 7.7     & 2.5    & 10.7       & 6.5      & 3.1      & 10.7      & 9.1      & 1.3     \\ 
\textsc{CapWAP}: \textsc{RL}       & 16.2     & 15.8     & 4.8     & 15.6     & 17.5     & 2.0    & 20.1       & 15.5      & 3.9      & 5.0      & 12.2      & 1.4     \\
\textsc{CapWAP}: \textsc{RL + SYN} & 13.9     & 11.1     & 4.8     & 13.8     & 8.7     & 2.0    & 13.6       & 7.6      & 4.4      & 9.6      & 7.0      & 2.6     \\ \bottomrule
\end{tabular}%
}
\caption{\emph{How do the generated captions differ qualitatively?} We present a number of automatic qualitative measures of caption content calculated over the dev sets: average caption length ($|\mathbf{y}|$), adjective production rate (adj\%), and the total vocabulary size of the unique unigrams and bigrams emitted ($|V^2|)$. Captions are measured in tokens (PTB-style), adjectives are identified using NLTK~\cite{nltk}, and vocabulary size is measured in thousands. Both \textsc{CapWAP} methods tend to produce longer captions, presumably with more descriptions (higher number of adjectives). Notably, \textsc{RL + SYN} manages to maintain more ``natural'' adjective production rates and richer language usage (in terms of bigram usage) than \textsc{RL} only, supporting the human quality ratings in Table~\ref{tab:syn-vs-rl}.}
\label{tab:qualitative}
\end{table*}

\begin{table*}[!t]
\centering
\resizebox{\textwidth}{!}{%
\renewcommand{\arraystretch}{1.2}
\begin{tabular}{@{}ll|ddd|ddd|ddd|ddd@{}}
\toprule
\multicolumn{1}{l}{\multirow{3}{*}{\textbf{Model}}} &
  \multirow{3}{*}{\textbf{Reward}} &
  \multicolumn{12}{c}{\textbf{Dataset}} \\
\multicolumn{1}{c}{} &
   &
  \multicolumn{3}{c}{CapVQA} &
  \multicolumn{3}{c}{CapGQA} &
  \multicolumn{3}{c}{CapVisual7W} &
  \multicolumn{3}{c}{CapVizWiz} \\
\multicolumn{1}{c}{} &
    &
  \multicolumn{1}{c}{IN} &
  \multicolumn{1}{c}{EM} &
  \multicolumn{1}{c}{F1} &
  \multicolumn{1}{c}{IN} &
  \multicolumn{1}{c}{EM} &
  \multicolumn{1}{c}{F1} &
  \multicolumn{1}{c}{IN} &
  \multicolumn{1}{c}{EM} &
  \multicolumn{1}{c}{F1} &
  \multicolumn{1}{c}{IN} &
  \multicolumn{1}{c}{EM} &
  \multicolumn{1}{c}{F1} \\ \midrule 
\textsc{RL + SYN}    & answer supported   & 42.5 & \mathbf{24}.\mathbf{7} & \mathbf{33}.\mathbf{8} & 25.1 & 16.9 & 20.2 & 16.7 & \mathbf{9}.\mathbf{1} & 15.2 & 34.3 & \mathbf{22}.\mathbf{0} & \mathbf{31}.\mathbf{3} \\
~~w/o ``no answer''  & answer most likely & 40.8 & 23.3 & 32.3 & 29.6 & \mathbf{19}.\mathbf{1} & \mathbf{23}.\mathbf{8} & 16.3 & 8.9 & 15.1 & 33.8 & 20.8 & 30.9 \\
~~w/o QA model       & answer present     & \mathbf{46}.\mathbf{4} & 22.4 & 31.5 & \mathbf{40}.\mathbf{2} & 13.9 & 21.6 & \mathbf{17}.\mathbf{4} & 8.9 & 15.2 & \mathbf{39}.\mathbf{0} & 13.5 & 26.6 \\ \midrule
Generic MLE          & None                        & 32.9 & 17.1 & 25.5 & 17.0 & 8.4 & 11.7 & 14.2 & 6.7 & 12.8 & 18.5 & 5.9 & 14.1 \\ \bottomrule
\end{tabular}%
}
\caption{\emph{What is the impact of using the QA model to provide rewards?} We present an ablation study across our different datasets when using the QA model with the ``no answer'' option or not, as well as a simple indicator reward, $\mathbf{1}\{\mathbf{a} \subseteq \mathbf{y}\}$, that simply measures if the answer string is present at all (without running the expensive QA model). Our results show that while the indicator reward increases the indicator metric (IN) the most, these are likely mostly spurious or disfluent generations. Using the QA model improves the F1 and EM scores across all datasets---and in all cases except one improves further when confidence is used.}
\label{tab:rc-abalation}
\end{table*}

\textbf{\mbox{Qualitative Discussion:}}
The qualitative effect of our method is quite intuitive (see Figure~\ref{fig:analysis-examples} as well as Table~\ref{tab:qualitative}). %, and Table~\ref{tab:qualitative} in Appendix~\ref{app:experiments}).
For example, many questions in CapGQA ask about spatial relations, which is reflected in the generated captions. On the other hand, CapVizWiz users often ask about detailed information about meals, and the adapted model attempts to provide a more useful description beyond a ``plate of food.'' 
Of the above datasets, note that only CapVizWiz consists of questions asked by genuinely interested users. Interestingly, this property unearths yet another challenge: CapVizWiz questions can be long-form and quite different from SQuAD, and reverse engineering them (using $F_\phi$) for pre-training is noisier (as evidenced by the performance of \textsc{+SYN} in Table~\ref{tab:syn-vs-rl}). While the artificial settings of the other datasets are not ideal, their diversity serves to demonstrate the flexibility of our approach.

\textbf{\mbox{Ablation Studies:}} Tables~\ref{tab:qualitative} and \ref{tab:rc-abalation} show the effects of different design choices in our \textsc{RL} and \textsc{RL + SYN} models. A significant challenge for \textsc{CapWAP} systems, as previously discussed and illustrated in Table~\ref{tab:fluency-and-fidelity}, is learning information need while maintaining fluency. Table~\ref{tab:qualitative} shows how synthetic pre-training regularizes the model to stay closer to human-level production patterns. Similarly, Table~\ref{tab:rc-abalation} shows how using the QA model to provide rewards (as opposed to a simple keyword search) helps the model avoid spurious rewards.

\textbf{Future Work:} The \textsc{CapWAP} paradigm introduces new challenges for learning effective systems, some of which our approach solves, and others which it still leaves open (e.g., maintaining fluency and fidelty). % while learning from weak supervision.
 While some may be addressed by large-scale multi-modal models~\cite{li2019visualbert,tan2019lxmert}, it is still unclear whether they would fully cover the diversity of information that real users are interested in (e.g., OCR). % such as textual information embedded within images (i.e., OCR).
%\footnote{Or even combinatorial needs, e.g., $\mathcal{D}_{\text{target}_1} \cup \mathcal{D}_{\text{target}_2}$.}

% The \textsc{CapWAP} paradigm introduces new challenges for learning effective systems---some of which our approach solves, and many others which it leaves open. Learning to maintain text fluency is one (Table~\ref{tab:fluency_and_fidelity}). Leveraging better image encoders (e.g., for OCR) is another. Finally, improvements to policy search are needed to learn to truly cater to richer information needs.
\section{Conclusion}
\label{sec:conclusion}

We defined and studied the $\textsc{CapWAP}$ task, where question-answer pairs provided by users are used as a source of supervision for learning their visual information needs. Our  results indicate that measuring caption content by its ability to logically support the answers to typical QA pairs from a target audience is (1) not only feasible, but also (2) a good proxy for uncovering information need. We hope this work will motivate the image captioning field to learn to anticipate and provide for the information needs of specific user communities.
% In order to learn an effective policy, our method builds upon standard RL by integrating a novel off-policy text planner---which provides  weakly supervised examples during pre-training. Evaluating the full system across multiple benchmarks showed the efficacy of our approach.

\section*{Acknowledgements}
We thank the MIT NLP group, the Google AI Language team, and the anonymous EMNLP reviewers for their valuable feedback. AF is supported in part by an NSF Graduate Research Fellowship.

% here we change the meaning of \VAN to use the prefix for the bibliography
\DeclareRobustCommand{\VAN}[3]{#3}
\bibliography{acl2020}
\bibliographystyle{acl_natbib}

\appendix
%\onecolumn
%auto-ignore
% \begin{center}
% {\bf \Large{Appendix}}
% \end{center}

\counterwithin{figure}{section}
\counterwithin{table}{section}

\section{\mbox{Model Architecture Details}}
\label{app:model}

The captioning architecture we use is a standard Transformer sequence-to-sequence model. As the model is not the main focus, we did not do any extensive hyper-parameter tuning or ablations beyond ensuring that we had a reasonable baseline model on COCO (114 CIDEr on COCO captions).

\textbf{Image Encoder:}
For each image $\mathbf{x}$, we take represent the region embeddings $\mathbf{o}_i \in \mathbb{R}^{2048}$ of the bounding boxes for the $k$ most confident object detections. We use the pre-trained Faster R-CNN~\cite{ren2015} model of \citet{Anderson2017up-down}.\footnote{\url{https://github.com/peteanderson80/bottom-up-attention}} We then map each region embedding to $\tilde{\mathbf{o}}_i \in \mathbb{R}^d$ using a single dense layer with a ReLU.
Inspired by the positional token embeddings in the BERT model~\cite{devlin-2019-bert}, we then augment $\tilde{\mathbf{o}}_i$ with learned position (the rasterized coordinate of the bounding box center), segment (a constant ``image'' component identifier), and confidence  (the detection rank of the object) embeddings to obtain the full object representation: $$\hat{\mathbf{o}}_i = \tilde{\mathbf{o}}_i + \mathbf{p}_i + \mathbf{s}_i + \mathbf{c}_i.$$

\textbf{Text Decoder:}  We decode the caption auto-regressively---starting with the \texttt{[CLS]} token and terminating on \texttt{[SEP]}. At each time-step $t$ we concatenate the image embeddings with special delimiters and the word-pieces decoded thus far, to obtain a joint context: $$\mathbf{h} = \{\texttt{[IMG]}, \hat{\mathbf{o}}_1, \ldots, \hat{\mathbf{o}}_k, \texttt{[CLS]}, \mathbf{w}_1, \ldots, \mathbf{w}_t\},$$ where $\mathbf{w}_i \in \mathbb{R}^d$ is the word piece embedding (the sum of token, position, and segment embeddings).
We then encode $\mathbf{h}$ using multi-layer Transformer, and compute the probability of generating $w_{t+1}$ using a softmax over the 30,522 word-piece vocabulary (the BERT vocabulary). For efficiency, we encode whole sequences at a time with a left-to-right attention mask: image regions may attend to all other image regions, and tokens may attend to all previous tokens and image regions.

\textbf{Hyperparameters:} In our experiments we use the top 64 object regions and a 6-layer Transformer with 512 hidden input units, 8 attention heads, and 2,048 hidden units in the intermediate feed-forward layer. During inference we do beam search with a beam size of 3 and  a length penalty $\alpha$ of $0.6$ \cite{wu2016google}. 

\textbf{Code:} We implemented our model in Tensorflow \cite{tensorflow2015-whitepaper}. Our code is available at \url{https://github.com/google-research/language/tree/master/language/capwap}.

\begin{figure*}[!t]
{\centering
\begin{minipage}{\linewidth}
\begin{algorithm}[H]
  \caption{ Synthetic data generation procedure for policy pre-training.}
    \label{alg:learn}
  \textbf{Definitions:} $\mathcal{D}_\text{generic}$ is assumed out-of-domain generic captioning data with input images $\tilde{\mathbf{x}}$ and supervised reference captions $\tilde{\mathbf{y}}$. $\mathcal{D}_\text{target}$ is the in-domain target QA data with input images $\mathbf{x}$, questions $\mathbf{q}$, and answers $\mathbf{a}$. $\mathcal{M}$ is the automatic QA model used in this paper for evaluation (Table~\ref{tab:task-definition}). $\mathcal{M}^{-1}(\mathbf{y})$ is a pre-trained QA generation model, that takes in some context $\mathbf{y}$ and outputs a predicted QA pair $(\hat{\mathbf{q}}, \hat{\mathbf{a}})$.
  \vspace{5pt}
 
  \begin{algorithmic}[1]
    \Function{train}{$\mathcal{D}_\text{generic}$, $\mathcal{M}^{-1}$, $T$}
    \Let{$\phi$}{random}\Comment{Initialize parameters for $F_\phi$}
      \For{$i = 1$ to  $T$}\Comment{  Train for $T$ steps}
        \State {$\tilde{\mathbf{x}}, \tilde{\mathbf{y}} \sim \mathcal{D}_\text{generic}$}\Comment{  Sample a generic image/caption pair}
        \Let{$\hat{\mathbf{q}}, \hat{\mathbf{a}}$}{$\mathcal{M}^{-1}(\mathbf{y})$}\Comment{Generate a synthetic QA pair}
        \Let{$\mathcal{L}$}{${-}\log F_\phi(\mathbf{y}|\mathbf{x}, \mathbf{q}, \mathbf{a})$}\Comment{Compute loss when conditioning on the QA}
        \Let{$\phi$}{$\Funct{minimize}{\mathcal{L}, \phi}$}\Comment{Update the model parameters of $F_\phi$}
      \EndFor
      \State \Return{$F_\phi$}\Comment{Produce $F_\phi$, the synthetic data generator}
    \EndFunction
    \\
    \Function{generate}{$\mathcal{D}_\text{target}$, $\mathcal{M}$, $F\phi$}
      \Let{$\mathcal{D}_\text{synthetic}$}{$[\,\,]$}\Comment{  Initialize synthetic dataset}
      \For{$(\mathbf{x}, \mathbf{q}, \mathbf{a}) \in \mathcal{D}_\text{target}$}\Comment{  Iterate target dataset}
        \Let{$\tilde{\mathbf{y}}$}{$\arg\!\max F_\phi(\mathbf{y} |\mathbf{x}, \mathbf{q}, \mathbf{a})$}\Comment{  Conditionally decode a caption for the QA pair}
         
         \If{$\mathcal{M}(\mathbf{q}, \tilde{\mathbf{y}}) = \mathbf{a}$}\Comment{  Filter for consistency}
            \State $\Funct{append}{\mathcal{D}_\text{synthetic}, (\mathbf{x}, \tilde{\mathbf{y}})}$\Comment{  Keep the synthetic sample}
        \EndIf
      \EndFor
      \State \Return{$\mathcal{D}_\text{synthetic}$}\Comment{ Yield $\mathcal{D}_\text{synthetic}$ for additional pre-training }
    \EndFunction
  \end{algorithmic}
\end{algorithm}
\end{minipage}
\par
}
\end{figure*}

\section{Model Training Details}
\label{app:reinforce}

\textbf{QA Model Threshold:}
During inference, the QA model $\mathcal{M}(\mathbf{q}, \mathbf{y})$ computes the probability of the ``no answer'' option  $p_{\mathcal{M}}(\texttt{NONE}|\mathbf{q}, \mathbf{y})$ and the probability of the most likely answer span $p_{\mathcal{M}}(\mathbf{y}_{i...j}|\mathbf{q}, \mathbf{y})$. We adjust how precise this reward is by treating the log odds ratio $c$ of the ``no answer'' vs. span options as a hyper-parameter when choosing the prediction $\hat{\mathbf{a}}$:

\begin{equation*}
  \log\left(\frac{p_{\mathcal{M}}(\mathbf{y}_{i...j}|\mathbf{q}, \mathbf{y})}{p_{\mathcal{M}}(\texttt{NONE}|\mathbf{q}, \mathbf{y})}\right) 
  \begin{cases}
  \begin{aligned}
  > c,  &  \hspace{0.5cm}\hat{\mathbf{a}}=\mathbf{y}_{i\ldots j} \\
  \leq c, & \hspace{0.5cm}\hat{\mathbf{a}}=\texttt{NONE}
  \end{aligned}
  \end{cases}
\end{equation*}

Depending on the value of $c$, we may only answer if we are confident the answer is supported---not just the most probable (e.g., based on answer-type)---to avoid potentially spurious rewards obtained by guessing or elimination. Table~\ref{tab:rc-abalation} shows an ablation over some choices of $c$.

\textbf{Answer Normalization:} For both training and evaluation, we normalize the gold and predicted answers by removing articles and punctuation when comparing them~\cite[see][]{rajpurkar-2016-squad}.

\textbf{Policy Gradient:}
We approximate the policy gradient (Eq.~\ref{eq:pg}) using a single Monte-Carlo sample $\mathbf{y} = (y_1, \ldots, y_n)$ from $ G_\theta(\mathbf{y}| \mathbf{x})$.
We accelerate training by restricting samples to be from a set of high-probability candidates with non-zero reward~\cite[cf.][\emph{inter alia}]{Anderson2017up-down, narayan-2018-ranking}. We decode using beam search and sample from the top-$k$ beams ($k=16$).

\textbf{Training:} For MLE pre-training on $\mathcal{D}_{\text{generic}}$ (Eq.~\ref{eq:cross-entropy}), all examples are shuffled and divided into mini-batches of 256 examples each. For RL adaptation to $\mathcal{D}_\text{target}$ (Eq.~\ref{eq:pg}), we use a mini-batch size of 128. To help regularize the fluency of the model, during RL training we continue to multi-task on the supervised generic captions, as in MIXER~\cite{Ranzato2015SequenceLT}. For both settings, we train for a maximum of 120K steps and choose the best model based on the dev set performance (using COCO CIDEr~\cite{VedantamZP15} for MLE pre-training and QA F1 for RL). For optimization, we use Adam \cite{kingma-2015-adam} with a linear warm-up and decay schedule.  Training was performed on a $4\times4$ TPU, and took about 1-2 hours per experiment.

\section{\mbox{Synthetic Pre-training Details}}
\label{app:synthetic}

\textbf{QA Conditional Model:} We use the same basic architecture for $F_\phi(\mathbf{y} | \mathbf{x}, \mathbf{q}, \mathbf{a})$ as for the main captioning model $G_\theta(\mathbf{y} | \mathbf{x})$, and only introduce two new ``question'' segment and ``answer'' segment embeddings that we add to differentiate the conditional text from the generated text in the Transformer. The full input then becomes:
\begin{align*}
\mathbf{h} = \{&\texttt{[IMG]}, \hat{\mathbf{o}}_1, \ldots, \hat{\mathbf{o}}_k, \texttt{[Q]}, \mathbf{q}_1, \ldots, \mathbf{q}_m, \\ &\texttt{[A]}, \textbf{a}_1, \ldots, \mathbf{a}_n, \texttt{[CLS]}, \mathbf{w}_1, \ldots, \mathbf{w}_t\},
\end{align*}
where the segment delimiter, $\mathbf{q}_i$, and $\mathbf{a}_j$ vectors are defined the same way as---and shared with---the caption's input word-piece embeddings $\mathbf{w}_i$ (see \S\ref{app:model}). We decode auto-regressively as before.

% \textbf{Question Generation Model:} We use the model of \citet{alberti-2019-synthetic} to generate synthetic questions $(\hat{\mathbf{q}}, \hat{\mathbf{a}})$ given contexts $\mathbf{y}$. At a high level, this inverse model
% (1) picks an answer span $\hat{\mathbf{a}} \subseteq \mathbf{y}$, (2) generates a question following some inferred distribution $p(\mathbf{q} | \mathbf{a}, \mathbf{y})$, and (3) confirms that the sample obeys ``round-trip filtering'', i.e., that the original QA model answers the synthetic example correctly ($\mathcal{M}(\hat{\mathbf{q}}, \mathbf{y}) = \hat{\mathbf{a}}$).
\label{app:datasets}

\begin{table*}[!t]
\begin{center}
\tabcolsep=0.4cm
\renewcommand{\arraystretch}{1.1}
\begin{tabular}{cl|cc|cc|cc}
\toprule
\multicolumn{2}{c|}{\bf Dataset} & \multicolumn{2}{c|}{\bf Train} & \multicolumn{2}{c|}{\bf Development} & \multicolumn{2}{c}{\bf Test}  \\
& & \text{Images} & \text{QA} & \text{Images} & \text{QA} & \text{Images} & QA\\
\midrule
\multirow{1}{*}{I} & \text{COCO} & $113,287$ & $-$ & $5,000$ & $-$ & $5,000$ & $-$ \\
\midrule
\multirow{4}{*}{II} & \text{CapVQA} & $104,311$ & $297,484$ & $4,617$ & $13,081$ & $4,615$ & $12,847$\\
&\text{CapGQA} & $69,450$ & $611,102$ & $5,000$ & $43,015$ & $4,739$ & $41,398$  \\
&\text{CapVisual7W} & $20,268$ & $93,878$  & $3,448$ & $16,314$ & $4,892$ & $22,769$  \\
&\text{CapVizWiz} & $10,027$ & $10,027$  & $960$ & $960$ & $1,905$ & $1,905$ \\
\bottomrule
\end{tabular}
\end{center}
\caption{\label{tab:data-stats} Statistics of the datasets used in this paper. Type I: generic/no QA. Type II: target/QA.}
\end{table*}

\textbf{Training and Generating:} 
We train $F_\phi(\mathbf{y} | \mathbf{x}, \mathbf{q}, \mathbf{a})$  using both the corpus of generic captions used for MLE pre-training in Section~\ref{sec:learning} (i.e., COCO) and additional Wikipedia text. We create automatic $(\mathbf{x}, \mathbf{q}, \mathbf{a}, \mathbf{y})$ and $(\texttt{null}, \mathbf{q}, \mathbf{a}, \mathbf{y})$ examples for COCO and Wikipedia sentences, respectively. To offset biases present in the question generation model (which is out-of-domain for caption-styled text as it is trained on SQuAD), we add $(\mathbf{x}, \texttt{null}, \mathbf{a}, \mathbf{y})$  examples from the generic captions by selecting random spans of $\mathbf{y}$, using the sampler of \citet{joshi2019spanBert} (random spans with Poisson distributed lengths). Algorithm~\ref{alg:learn} illustrates the full synthetic  data generation process.

\section{Converted Dataset Details}

\textbf{Splits:} For the COCO and VQA datasets we use the `Karpathy' splits from \citet{karpathy}. For GQA, we use the `balanced' splits, but limit to $\sim$5K images each for the new test and dev sets from the original GQA dev set. Both the Visual7W and GQA datasets have images from Visual Genome~\cite{krishnavisualgenome}, and thus some (partially) overlap with the `Karpathy' COCO images. Since we use COCO for pre-training ($\mathcal{D}_\text{generic}$), we avoid data leakage by mapping Visual Genome IDs to COCO IDs and either filter questions about images that are in the COCO train or dev sets, or re-assign the data to match COCO. Finally, on VizWiz, we combine all of the original data and randomly re-partition it into test and dev sets of $\sim$1K and $\sim$2K images each,  keeping the rest for training.

\end{document}